%% file: main.tex
\documentclass[letterpaper]{article} 
\usepackage[preprint]{aaai2027} 
\usepackage[hyphens]{url} 
\usepackage{graphicx} 
\usepackage{natbib} 
\usepackage{caption} 
\usepackage{amsmath}
\usepackage{amssymb}
\usepackage{booktabs}
\usepackage{multirow}
\usepackage{colortbl}
\usepackage{tikz}
\definecolor{configband}{RGB}{234,248,247}
\definecolor{availpresent}{RGB}{180,180,180}
\tikzset{
  availability/base/.style={
    circle,
    minimum size=3.7mm,
    inner sep=0pt,
    outer sep=0pt,
    line width=0.28pt,
    font=\fontsize{5.2}{5.2}\selectfont
  },
  availability/present/.style={
    availability/base,
    draw=availpresent,
    fill=availpresent
  },
  availability/missing/.style={
    availability/base,
    draw=black,
    fill=white,
    dash pattern=on 1.2pt off 0.8pt
  }
}
\newcommand{\AvailNode}[4]{%
  \ifnum#1=1\relax
    \node[availability/present] at (#2,#3) {#4};
  \else
    \node[availability/missing] at (#2,#3) {#4};
  \fi
}
\newcommand{\AvailConfig}[4]{%
  \begin{tikzpicture}[
    baseline=(current bounding box.center),
    x=4mm,
    y=4mm
  ]
    \AvailNode{#1}{0}{1}{F}
    \AvailNode{#2}{1}{1}{T1}
    \AvailNode{#3}{0}{0}{T1\raisebox{-0.2ex}{\scalebox{0.75}{c}}}
    \AvailNode{#4}{1}{0}{T2}
  \end{tikzpicture}%
}
\newcommand{\AvailHeader}[4]{\cellcolor{white}\AvailConfig{#1}{#2}{#3}{#4}}

\title{LASSNet: Level-Aware Availability-Conditioned Spatial-Semantic Fusion for Brain Tumor Segmentation with Missing MRI Modalities}
\author{
    Haobin Chen\textsuperscript{\rm 1,\rm 2},
    Ao Chang\textsuperscript{\rm 1,\rm 2},
    Rundong Wang\textsuperscript{\rm 1,\rm 2},
    Zhicheng Li\textsuperscript{\rm 1,\rm 2},
    Zhihao Tang\textsuperscript{\rm 3},
    Heqin Zhu\textsuperscript{\rm 1,\rm 2}\corresponding
}
\affiliations{
    \textsuperscript{\rm 1}School of Biomedical Engineering, Division of Life Sciences and Medicine, University of Science and Technology of China (USTC), Hefei, Anhui, 230026, China\\
    \textsuperscript{\rm 2}Medical Imaging, Robotics, Analytic Computing \& Learning (MIRACLE) Lab, YRD-RIGHT, Suzhou Institute for Advanced Research, USTC, Suzhou, Jiangsu, 215123, China\\
    \textsuperscript{\rm 3}College of Information Science and Technology, Donghua University\\
    zhuheqin@ustc.edu
}

\begin{document}

\maketitle

\begin{abstract}
Brain tumor segmentation from multimodal MRI relies on complementary evidence across four imaging sequences, yet one or more modalities may be unavailable because of acquisition cost, protocol variation, scan failure, or patient condition. Existing work has explored reconstruction, knowledge transfer, and direct feature fusion, but leaves open whether missing-modality fusion should change with representation level. High-resolution lateral features retain spatial detail, whereas compressed bottleneck features encode semantic and inter-modality context. We therefore hypothesize that fusion should be conditioned jointly on modality availability and feature hierarchy. We propose the Level-Aware Availability-Conditioned Spatial-Semantic Fusion Network (LASSNet), which contains two level-specialized modules. Hierarchical Availability-Conditioned Fusion (HACF) constructs four lateral representations using count-normalized aggregation of available modalities, mask-conditioned channel modulation, and local 3D refinement. Tri-Scale Relational--Spatial Fusion (TriRSF) models relations among available modality descriptors and spatial context across multiple bottleneck resolutions, followed by cross-scale aggregation and availability-conditioned global spatial attention. A shared coarse-to-fine decoder starts from TriRSF semantics and progressively injects HACF features, without reconstructing missing inputs. Across all 15 non-empty modality configurations, LASSNet obtains mean Dice scores of 76.7\% and 83.2\% over WT, TC, and ET on BraTS2019 and BraTS2023, respectively.
\end{abstract}

\section{Introduction}

Accurate brain tumor delineation supports quantitative assessment and treatment planning~\cite{chang2025p2ed,chang2024pemed}. Multimodal magnetic resonance imaging (MRI) combines complementary evidence from FLAIR, T1, contrast-enhanced T1 (T1ce), and T2 sequences~\cite{menze2015brats,bakas2017advancing}. However, complete-input models are difficult to deploy when sequences are absent because of acquisition cost, protocol variation, artifacts, contraindications, or patient motion, motivating segmentation from any non-empty modality subset~\cite{zhou2023survey}.

Existing methods reconstruct absent data, transfer knowledge from complete observations, or directly fuse available representations~\cite{dorent2019uhved,wang2021acn,yang2022d2net}. Direct fusion avoids dependence on synthesized inputs and has progressed from feature statistics~\cite{havaei2016hemis} to gated convolution~\cite{chen2019robust,ding2021rfnet}, shared--specific representations~\cite{wang2023shaspec}, and Transformer or state-space fusion~\cite{zhang2022mmformer,shi2023mftrans,pipoli2025imfuse}.

Most advances determine \emph{which} observed modalities contribute and \emph{how} to combine them, but leave less explicit whether fusion should vary with representation level. High-resolution lateral features restore boundaries and small structures, whereas bottleneck features encode semantic, inter-modality, and long-range context. Missingness also perturbs these hierarchies differently. In lateral paths, modality count changes feature magnitude, while subset identity changes which channels remain reliable for boundary recovery. At the bottleneck, spatial compression makes explicit reasoning over modality complementarity and long-range tumor context both more useful and computationally affordable. This creates a level-aware problem chain: normalize variable-cardinality evidence, adapt detail features to the observed subset, and model semantic relations at compact scales. Applying one fusion behavior throughout risks either under-modeling bottleneck complementarity or imposing unnecessarily heavy relation modeling on high-resolution skips.

We address this distinction with two availability-conditioned paths. Hierarchical Availability-Conditioned Fusion (HACF) forms four lateral features through count-normalized aggregation, mask-conditioned channel modulation, and local 3D refinement. Tri-Scale Relational--Spatial Fusion (TriRSF) models available-modality relations and spatial context at the bottleneck across three resolutions. The paths remain parallel until decoding, preventing transformed bottleneck semantics from overwriting the high-resolution evidence they should complement. TriRSF initializes a shared decoder that progressively incorporates HACF detail, enabling direct segmentation without synthesizing missing sequences.

Our contributions are threefold:
\begin{itemize}
    \item We formulate fusion jointly over modality availability and representation level, and introduce HACF for count-stable, configuration-conditioned lateral aggregation.
    \item We develop TriRSF to couple availability-masked modality relations with multi-scale spatial modeling, and connect it to HACF through coarse-to-fine decoding.
    \item We evaluate one model over all 15 non-empty modality configurations on BraTS2019 and BraTS2023, obtaining mean WT/TC/ET Dice scores of 76.7\% and 83.2\%.
\end{itemize}

\section{Related Work}

\paragraph{Reconstruction and representation transfer.}
Missing-modality methods either synthesize unavailable evidence or constrain incomplete representations using complete observations. U-HVED learns a variational completion space~\cite{dorent2019uhved}, while M$^3$AE combines masked autoencoding, model inversion, and self-distillation~\cite{liu2023m3ae}; subsequent approaches reconstruct features, personalize all-stage completion, or use diffusion-based auxiliaries~\cite{diao2023joint,zhao2024allstage,zhao2025adaptive,ali2026d3seg}. Knowledge-transfer methods employ adversarial co-training, teacher selection, modality-dependent distillation, or semantic-relation transfer~\cite{wang2021acn,zeng2024deconet,wang2023lckd,wang2024metakd,su2024rirn,shi2024passion}, whereas ShaSpec, MedMAP, KMD, and DC-Seg regularize shared, aligned, invariant, or anatomy-aware representations~\cite{wang2023shaspec,liu2026medmap,liu2025kmd,li2025dcseg}. These strategies exploit complete training data but introduce generation, teacher, or alignment objectives; LASSNet instead focuses on direct availability-conditioned fusion.

\paragraph{Direct latent-space fusion.}
Direct methods aggregate observed features without requiring synthesized inputs. HeMIS uses feature statistics~\cite{havaei2016hemis}; RobustSeg, RFNet, and SFusion introduce gated, region-aware, or self-attentive fusion~\cite{chen2019robust,ding2021rfnet,liu2023sfusion}; and Transformer variants model modality interactions through availability-aware tokens, multi-scale features, or token merging~\cite{zhang2022mmformer,shi2023mftrans,qiu2024mmmvit,zhang2024tmformer}. Recent state-space approaches include Mamba fusion at lateral and bottleneck sites, learned modality ordering, and contribution calibration~\cite{pipoli2025imfuse,zhang2025ls3m,qiu2025pndc,zou2026mgml}. These methods improve how observed modalities interact, but commonly apply one fusion family at a selected site or repeatedly across stages. Our question is narrower: whether the fusion operator itself should change between lateral spatial detail and bottleneck semantics.

\paragraph{Level-aware spatial--semantic fusion.}
Volumetric encoder--decoders combine compact semantics with multi-resolution lateral detail~\cite{cicek20163dunet,chang2024emnet}, while self-attention supports non-local spatial interaction~\cite{vaswani2017attention}. Hierarchical In-Out Fusion likewise distinguishes network levels, using shallow self-attention and deepest-layer spatial--frequency fusion after intra-modal encoding~\cite{liu2025inoutfusion}. LASSNet differs by making modality availability explicit in two complementary paths: HACF performs count-normalized, mask-conditioned lateral fusion and local refinement, whereas TriRSF models available-modality relations and spatial context across three bottleneck scales. Its contribution is therefore the availability-conditioned, level-specific organization and decoder coupling of these established operations.

\input{tables/brats_per_config_tables.tex}
\input{method.tex}

\FloatBarrier

\section{Experiments}
\label{sec:experiments}

\input{tables/fusion_ablation_table.tex}

\paragraph{Implementation Details.}
We crop each case to the nonzero brain bounding box, apply foreground z-score normalization, remap label 4 to label 3, and train on random $128^3$ crops with rotation, intensity perturbation, and flips. PyTorch runs use one NVIDIA RTX 4090, batch size one, and 150 iterations per epoch. RAdam starts at $2\times10^{-4}$ with weight decay $3\times10^{-5}$ and schedule $(1-e/E)^{0.9}$. Each sample uniformly draws one of the 15 non-empty availability masks and uses Eq.~\ref{eq:objective}. Patient-disjoint train/validation/test splits contain 234/33/68 BraTS2019 cases~\cite{bakas2018brats} and 875/125/251 BraTS2023 volumes~\cite{baid2024brats2023}. BraTS2019 is trained from scratch for 1000 epochs and evaluated at the terminal checkpoint. BraTS2023 continues a checkpoint of the same architecture without the ET-specific term through epoch 1680 and selects the best validation checkpoint; validation occurs every 20 epochs on the four single-modality and complete inputs. Testing covers all 15 masks with $128^3$ sliding windows at 50\% overlap. Dice is averaged over subjects and then masks for WT (labels 1--3), TC (1 and 3), and ET (3).

\paragraph{Comparison with Existing Methods.}
We compare LASSNet with U-HVED~\cite{dorent2019uhved}, RobustSeg~\cite{chen2019robust}, SFusion~\cite{liu2023sfusion}, mmFormer~\cite{zhang2022mmformer}, ShaSpec~\cite{wang2023shaspec}, M$^3$AE~\cite{liu2023m3ae}, M$^3$FeCon~\cite{zeng2024m3fecon}, and IM-Fuse~\cite{pipoli2025imfuse}. Tables~\ref{tab:brats2019} and~\ref{tab:brats2023} align all results in the same 15-configuration order and use region-wise macro averages. \emph{BraTS2019.} LASSNet obtains 61.9\%, 80.4\%, and 87.9\% Dice for ET, TC, and WT, yielding 76.7\% overall. It exceeds the strongest comparison, mmFormer, by 1.4, 0.6, 0.5, and 0.8 points, respectively.
The configuration-wise results follow the expected contrast roles rather than being driven only by complete input. ET rises from 49.3\% with FLAIR, T1, and T2 to 76.0\% with T1, T1ce, and T2, whereas WT is already strong with FLAIR alone (88.9\%) or T2 alone (87.8\%). Thus, LASSNet preserves fluid-sensitive evidence for tumor extent while exploiting T1ce for enhancing tissue.

\emph{BraTS2023.} LASSNet reaches 73.6\% ET, 85.5\% TC, and 90.5\% WT, or 83.2\% overall. Against IM-Fuse, it leads WT by 0.3 points, ties TC, and trails ET and the overall mean by 0.7 and 0.1 points. The same regional pattern persists: FLAIR-only WT reaches 91.9\%, while adding T1ce to the FLAIR--T1--T2 subset raises ET from 65.1\% to 83.6\%. The near tie in overall mean therefore masks a regional trade-off: LASSNet is stronger on spatially extensive WT, while small enhancing tissue remains its clearest optimization target. Thus, the leading comparison changes across benchmark versions, while LASSNet remains competitive on both datasets.

\paragraph{Ablation Study and Visualization.}
Table~\ref{tab:ablation} compares masked-mean fusion, uniform availability-conditioned fusion, either specialized path alone, and Full LASSNet on BraTS2019. Under the same 1000-epoch schedule, Full obtains 87.89\%, 80.38\%, and 61.88\% Dice for WT, TC, and ET, raising Mean Dice from 73.84\% for Base to 76.72\%. Neither specialized path alone surpasses Base (73.76\% without HACF and 73.43\% without TriRSF), whereas their joint configuration gains 2.88 points. This non-additive result indicates that the principal benefit comes from coupling configuration-adapted lateral detail with relational bottleneck semantics, rather than inserting either module in isolation. The 4.87-point ET gain is the largest, suggesting that the coupling is particularly useful for the smallest, availability-sensitive region, although repeated runs are needed to quantify stability. Figure~\ref{fig:qualitative} visualizes LASSNet segmentation results under all 15 missing-modality configurations for a BraTS2023 test case using a fixed slice and crop. Consistent with Table~\ref{tab:brats2023}, removing T1ce from the complete input reduces average ET Dice from 83.6\% to 65.1\%.

\section{Discussion and Conclusion}

LASSNet encodes the inductive bias that lateral features should recover spatial detail while bottleneck features should model available-modality relations and semantic context, rather than applying one fusion operator uniformly across representation levels. HACF and TriRSF instantiate these complementary roles and reconnect through a shared coarse-to-fine decoder without reconstructing missing inputs. Across all 15 non-empty configurations, LASSNet obtains mean Dice scores of 76.7\% and 83.2\% on BraTS2019 and BraTS2023. Under the matched 1000-epoch schedule, the Full configuration improves Mean Dice over Base by 2.88 points, with the largest gain on ET, supporting the joint use of both fusion paths. These results establish availability-conditioned level specialization as a coherent approach to incomplete multimodal segmentation. Future work will examine repeated-run stability, naturally incomplete multi-center cohorts, uncertainty-aware prediction, and computational efficiency.

\FloatBarrier
\bibliography{aaai2027}

\end{document}

%% file: tables/brats_per_config_tables.tex
\newcommand{\PerConfigResultsTableBlock}[3]{%
\centering
\begingroup
\small
\setlength{\tabcolsep}{0.75pt}
\renewcommand{\arraystretch}{0.88}
\begin{tabular*}{\textwidth}{@{\extracolsep{\fill}}c l
>{\columncolor{configband}}c c
>{\columncolor{configband}}c c
>{\columncolor{configband}}c c
>{\columncolor{configband}}c c
>{\columncolor{configband}}c c
>{\columncolor{configband}}c c
>{\columncolor{configband}}c c
>{\columncolor{configband}}c c@{}}
\toprule
& \textbf{Model}
& \AvailHeader{1}{0}{0}{0}
& \AvailHeader{0}{1}{0}{0}
& \AvailHeader{0}{0}{1}{0}
& \AvailHeader{0}{0}{0}{1}
& \AvailHeader{1}{1}{0}{0}
& \AvailHeader{1}{0}{1}{0}
& \AvailHeader{1}{0}{0}{1}
& \AvailHeader{0}{1}{1}{0}
& \AvailHeader{0}{1}{0}{1}
& \AvailHeader{0}{0}{1}{1}
& \AvailHeader{1}{1}{1}{0}
& \AvailHeader{1}{1}{0}{1}
& \AvailHeader{1}{0}{1}{1}
& \AvailHeader{0}{1}{1}{1}
& \AvailHeader{1}{1}{1}{1}
& \cellcolor{white}\textbf{Avg.} \\
\midrule
\input{#3}%
\end{tabular*}
\endgroup
\caption{#1}
\label{#2}
}

%% file: method.tex
\section{Method}
\label{sec:method}

\subsection{Problem Formulation and Framework Overview}
\label{subsec:overview}

Incomplete multimodal segmentation requires one model to handle every non-empty subset of the imaging sequences.
Because high-resolution lateral features preserve spatial detail while bottleneck features encode semantic and cross-modality context, LASSNet assigns availability-conditioned fusion operators according to representation level.

Let $\mathcal{M}=\{\mathrm{FLAIR},\mathrm{T1ce},\mathrm{T1},\mathrm{T2}\}$ contain $M=4$ modalities, with co-registered input $\mathbf{X}=\{\mathbf{X}_m\}_{m=1}^{M}$ and availability vector $\mathbf{a}\in\{0,1\}^{M}$, where $a_m=1$ denotes an observed modality and $\sum_m a_m\geq1$.
We learn
\begin{equation}
\mathbf{p}=f_{\theta}(\mathbf{X},\mathbf{a})\in[0,1]^{C\times H\times W\times D},
\qquad
\sum_{c=1}^{C}p_{c,v}=1,
\label{eq:problem}
\end{equation}
where $\mathbf{p}$ is the voxel-wise posterior over $C$ classes and $v$ indexes a voxel.

\begin{figure*}[!t]
\centering
\includegraphics[width=\textwidth]{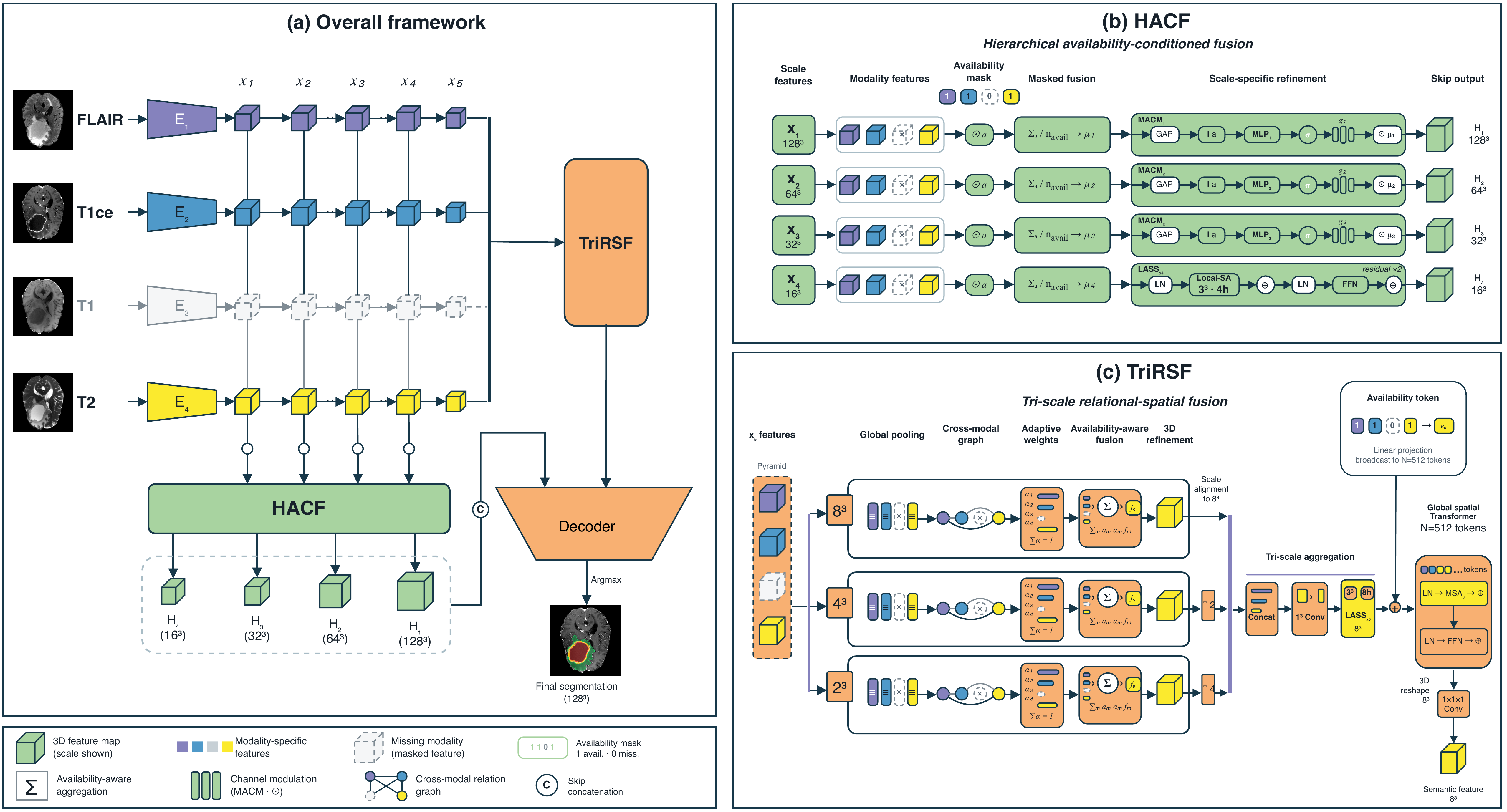}
\caption{Overview of the proposed framework. HACF constructs availability-conditioned lateral representations, while TriRSF models modality relations and spatial context at the bottleneck. They are parallel paths coupled by coarse-to-fine decoding; TriRSF does not take HACF as its input. The final global Transformer operates over spatial tokens after modality aggregation.}
\label{fig:framework}
\end{figure*}

As shown in Fig.~\ref{fig:framework}, four modality-specific encoders feed two parallel paths: HACF constructs availability-conditioned lateral features, while TriRSF models modality relations and spatial context at the bottleneck.
A shared decoder starts from TriRSF semantics, progressively injects HACF detail, and passes its prediction to an availability-conditioned refinement head.
Keeping these paths separate is intentional: HACF transmits spatially aligned evidence, whereas TriRSF transforms compressed semantics more aggressively. The availability vector therefore conditions distinct decisions: channel reliability on lateral grids, and node participation plus spatial context at the bottleneck. This division avoids forcing one operator to satisfy conflicting detail-preservation and context-modeling requirements.
LASSNet directly segments observed inputs without reconstructing missing sequences.

\subsection{Modality-Specific Hierarchical Representation Learning}
\label{subsec:encoding}

\paragraph{Five-level modality-specific encoder.}
Following 3D U-Net~\cite{cicek20163dunet}, each modality is processed by an architecturally identical but parameter-independent five-stage residual encoder $E_m$.
The first stage preserves resolution and each subsequent stage downsamples by two, producing
\begin{equation}
\{\mathbf{F}_m^\ell\}_{\ell=1}^{5}=E_m(\mathbf{X}_m),
\qquad
\mathbf{F}_m^\ell\in\mathbb{R}^{C_\ell\times S_\ell\times S_\ell\times S_\ell},
\label{eq:encoder}
\end{equation}
Channels are $(8,16,32,64,128)$ and grid sizes are $(128,64,32,16,8)$.
Pre-normalized 3D convolutions refine each stage without mixing modalities.

\paragraph{Intra-modal global context encoding.}
The fifth-level feature is projected to 512 channels, flattened into $N=512$ spatial tokens of dimension $d=512$, and augmented with a modality-specific positional embedding:
\begin{equation}
\mathbf{Q}_m=\operatorname{Flat}\!\left(P_m(\mathbf{F}_m^5)\right)+\mathbf{P}_m,
\qquad
\mathbf{Q}_m\in\mathbb{R}^{N\times d}.
\label{eq:tokenization}
\end{equation}
An independent one-layer Transformer~\cite{vaswani2017attention} then models intra-modal spatial dependencies:
\begin{equation}
\begin{aligned}
\mathbf{U}_m&=\mathbf{Q}_m+
\operatorname{MSA}_m\!\left(\operatorname{LN}(\mathbf{Q}_m)\right),\\
\mathbf{T}_m&=\mathbf{U}_m+
\operatorname{FFN}_m\!\left(\operatorname{LN}(\mathbf{U}_m)\right).
\end{aligned}
\label{eq:intra_transformer}
\end{equation}
It uses eight attention heads and a 4096-dimensional feed-forward expansion.
Because encoding precedes fusion, tokens attend only within one modality before being reshaped as $\mathbf{Z}_{m,8}\in\mathbb{R}^{512\times8\times8\times8}$.

\paragraph{Availability-aware feature masking.}
At each fusion interface, the explicit availability vector removes unavailable branches from a stacked feature $\mathbf{Y}$:
\begin{equation}
\widetilde{\mathbf{Y}}_m=a_m\mathbf{Y}_m.
\label{eq:feature_mask}
\end{equation}
Thus, missing branches contribute to neither fusion path, and one parameter set supports all $2^M-1=15$ configurations.

\input{tables/brats2019_per_config_table.tex}
\input{tables/brats2023_per_config_table.tex}

\subsection{Hierarchical Availability-Conditioned Fusion}
\label{subsec:hacf}

\paragraph{HACF for spatial detail preservation.}
HACF preserves lateral spatial detail while stabilizing feature scale across modality subsets.
At level $\ell\in\{1,2,3,4\}$, it averages only observed features:
\begin{equation}
\overline{\mathbf{F}}^\ell=
\frac{\sum_{m=1}^{M}a_m\mathbf{F}_m^\ell}
{\max\!\left(\sum_{m=1}^{M}a_m,\epsilon\right)}.
\label{eq:masked_mean}
\end{equation}
Count normalization yields a fixed-width representation with comparable magnitude for any number of available modalities.
The masked mean is an anchor: it removes modality cardinality as a nuisance factor before learned components model subset- and content-dependent corrections. Otherwise, the gate must learn magnitude normalization and semantic reweighting simultaneously.

\paragraph{Mask-conditioned channel modulation.}
At the first three levels, a two-layer MLP maps the pooled mean and complete availability vector to a sigmoid channel gate:
\begin{equation}
\begin{aligned}
\mathbf{g}^{\ell}
&=\sigma\!\left(
W_{2}^{\ell}\phi\!\left(
W_{1}^{\ell}
\left[\operatorname{GAP}(\overline{\mathbf{F}}^\ell);\mathbf{a}\right]
\right)\right),\\
\mathbf{H}^{\ell}
&=\overline{\mathbf{F}}^\ell\odot\mathbf{g}^{\ell},
\qquad \ell\in\{1,2,3\},
\end{aligned}
\label{eq:hacf_modulation}
\end{equation}
where $\phi$ is ReLU.
The descriptor captures sample content, while $\mathbf{a}$ identifies its contributing contrasts, allowing subset-specific channel reweighting without changing the spatial grid.
Modulation follows aggregation rather than acting independently on each stream, so its cost is independent of the number of observed modalities and the decoder always receives the same feature geometry. At the three densest levels, restricting adaptation to channels also avoids expensive voxel-wise relation modeling that could disturb boundary coordinates.

\paragraph{Local spatial consistency refinement.}
At the fourth level, local 3D attention and a spatial feed-forward network refine the $16^3$ representation:
\begin{equation}
\begin{aligned}
\mathbf{V}^{4}
&=\overline{\mathbf{F}}^{4}
+\gamma_{4,1}\mathcal{A}_{\mathrm{loc}}
\!\left(\operatorname{LN}_{c}(\overline{\mathbf{F}}^{4})\right),\\
\mathbf{H}^{4}
&=\mathbf{V}^{4}
+\gamma_{4,2}\mathcal{F}_{\mathrm{sp}}
\!\left(\operatorname{LN}_{c}(\mathbf{V}^{4})\right).
\end{aligned}
\label{eq:hacf_local}
\end{equation}
Here, $\mathcal{A}_{\mathrm{loc}}$ uses four heads within $3^3$ neighborhoods, and $\mathcal{F}_{\mathrm{sp}}$ combines channel expansion, GELU, depthwise $3^3$ convolution, and projection.
Zero-initialized residual scales preserve the initial identity mapping.
The fourth level is a transition point: its grid is compact enough for neighborhood attention but remains directly aligned with a decoder skip. Local refinement can therefore repair spatial inconsistencies introduced by averaging without importing global bottleneck transformations.
HACF returns $\{\mathbf{H}^{\ell}\}_{\ell=1}^{4}$ without cross-scale interaction inside the skip path.
This omission is deliberate; every skip retains its native decoder alignment, while cross-scale semantic interaction is delegated to TriRSF.

\subsection{TriRSF: Tri-Scale Relational--Spatial Fusion}
\label{subsec:trirsf}

\paragraph{Tri-scale bottleneck pyramid.}
Unlike the lateral path, bottleneck fusion must capture semantic complementarity among the observed contrasts.
TriRSF models modality importance, inter-modality relations, and spatial context at this compact level.
Two shared stride-two 3D convolutions form $\mathbf{Z}_{m,4}=D_1(\mathbf{Z}_{m,8})$ and $\mathbf{Z}_{m,2}=D_2(\mathbf{Z}_{m,4})$, producing a 512-channel pyramid at $8^3$, $4^3$, and $2^3$ with progressively larger effective receptive fields.
These scales are complementary rather than duplicate predictions: the native grid retains localized tumor patterns, while the downsampled grids expose coarser context and make descriptor relations less sensitive to voxel-level noise. Shared downsampling keeps modality features in comparable channel spaces.

\paragraph{Availability-conditioned modality relation modeling.}
At scale $s\in\{8,4,2\}$, global average pooling yields descriptor $\mathbf{d}_{m,s}$.
A scoring network combines it with $\mathbf{a}$ to produce availability-masked weights and a base mixture:
\begin{equation}
\begin{aligned}
e_{m,s}&=q_s([\mathbf{d}_{m,s};\mathbf{a}]),\\
\alpha_{m,s}&=
\frac{a_m\exp(e_{m,s})}{\sum_j a_j\exp(e_{j,s})},
\qquad
\overline{\mathbf{Z}}_s=\sum_m\alpha_{m,s}\mathbf{Z}_{m,s}.
\end{aligned}
\label{eq:base_weights}
\end{equation}
Thus, unavailable modalities receive zero mass, while observed-modality weights depend on both sample content and subset identity.
This base mixture provides a stable first-order estimate of modality importance. The subsequent graph asks a distinct conditional question: how should one modality's contribution change after observing the descriptors of all other surviving contrasts?
To model dependencies beyond independent importance, descriptors are projected into graph nodes $\mathbf{n}_{m,s}$ and connected only when available.
Scaled dot-product edges guide two residual message-passing steps:
\begin{equation}
\begin{aligned}
A_{mj,s}&=
\operatorname{Softmax}_{j:a_j=1}
\left(
\frac{\mathbf{n}_{m,s}^{\top}\mathbf{n}_{j,s}}{\sqrt{d_g}}
\right),\qquad
\mathbf{h}_{m,s}^{0}=\mathbf{n}_{m,s},\\
\mathbf{h}_{m,s}^{t+1}
&=\mathbf{h}_{m,s}^{t}
+\psi_s\!\left(
\sum_j A_{mj,s}\mathbf{h}_{j,s}^{t}
\right),
\qquad t\in\{0,1\}.
\end{aligned}
\label{eq:relation_graph}
\end{equation}
The updated nodes and $\mathbf{a}$ produce graph-aware masked weights $\beta_{m,s}$ and mixture $\mathbf{Z}^{\mathrm{rel}}_s=\sum_m\beta_{m,s}\mathbf{Z}_{m,s}$.
Descriptor-level nodes avoid quadratic attention over all modality--voxel pairs and support missingness naturally by excluding unavailable nodes from normalization. Two residual propagation steps expose each node to joint subset context while retaining its own descriptor.
A zero-gated residual projection combines both mixtures:
\begin{equation}
\mathbf{R}_s=\overline{\mathbf{Z}}_s+
\gamma_s P_s[\overline{\mathbf{Z}}_s;\mathbf{Z}^{\mathrm{rel}}_s].
\label{eq:relational_fusion}
\end{equation}
Initializing $\gamma_s=0$ preserves the base fusion initially, and using one node per modality keeps relation modeling compact.
Accordingly, relational fusion is learned as a correction to a valid base estimate rather than imposed as its replacement, which stabilizes optimization when simple aggregation is already sufficient for a subset.

\paragraph{Relational--spatial refinement and scale aggregation.}
Independent local 3D attention--feed-forward blocks produce $\mathbf{L}_s$.
The coarser features are upsampled to $8^3$, concatenated, projected, and locally consolidated:
\begin{equation}
\mathbf{V}=
P_{\mathrm{tri}}
\left[
\mathbf{L}_8;
\uparrow_{8}\mathbf{L}_4;
\uparrow_{8}\mathbf{L}_2
\right],
\qquad
\widehat{\mathbf{V}}=\mathcal{R}_{\mathrm{loc}}(\mathbf{V}).
\label{eq:tri_scale}
\end{equation}
The projection restores 512 channels, combining fine bottleneck structure with coarse semantic context.
Local refinement precedes scale aggregation so that neighborhood structure is resolved at each native receptive field. Global attention is then applied only once after aggregation, separating local geometric coherence from long-range dependency modeling and avoiding repeated global attention at all three scales.
Finally, $\widehat{\mathbf{V}}$ is flattened into spatial tokens, augmented with broadcast availability and positional embeddings, and processed by a global Transformer:
\begin{equation}
\begin{aligned}
\mathbf{Q}_{f}=
\operatorname{Flat}(\widehat{\mathbf{V}})
+\mathbf{1}_{N}(W_a\mathbf{a})^\top
+\mathbf{P}_{f},\\
\mathbf{G}=
P_G\operatorname{Reshape}
\left(
\mathcal{T}_{\mathrm{global}}(\mathbf{Q}_{f})
\right).
\end{aligned}
\label{eq:global_transformer}
\end{equation}
Because modality aggregation is complete, this attention operates over spatial rather than modality tokens, while the availability embedding records the contributing subset.
Retaining this embedding is important because subset identity remains informative after the explicit modality dimension has disappeared.
The resulting $\mathbf{G}\in\mathbb{R}^{512\times8\times8\times8}$ initializes the decoder.

\begin{figure*}[!t]
\centering
\includegraphics[width=\textwidth]{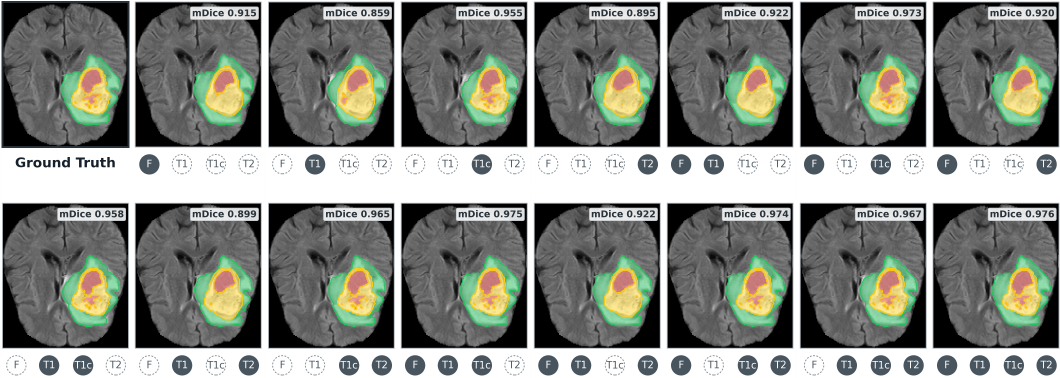}
\caption{Visualization of LASSNet segmentation results under missing modality scenarios. Enhancing tumor (yellow), necrotic tissue (red), and edema (green).}
\label{fig:qualitative}
\end{figure*}

\subsection{Coarse-to-Fine Decoding and Error-Guided Refinement}
\label{subsec:decoder}

\paragraph{Shared decoder.}
The shared decoder first couples both paths by compressing $\mathbf{G}$ to 128 channels and concatenating each upsampled semantic feature with the corresponding HACF representation:
\begin{equation}
\begin{aligned}
\mathbf{D}^{5}&=\Phi_{5}(\mathbf{G}),\\
\mathbf{D}^{\ell}
&=\Phi_{\ell}
\left[
\uparrow\mathbf{D}^{\ell+1};
\Psi_{\ell}(\mathbf{H}^{\ell})
\right],
\qquad \ell\in\{4,3,2,1\}.
\end{aligned}
\label{eq:decoder}
\end{equation}
Four intermediate heads provide full-resolution deep supervision, and a final $1\times1\times1$ classifier produces $\mathbf{p}^{\mathrm{base}}$.

\paragraph{Availability-conditioned posterior refinement.}
To correct subset-dependent uncertainty, the refinement head combines the base posterior, highest-resolution HACF feature, and broadcast availability vector:
\begin{equation}
\mathbf{h}=
\rho\!\left(
\left[
\mathbf{p}^{\mathrm{base}};
\mathbf{H}^{1};
\operatorname{Tile}(\mathbf{a})
\right]
\right),
\label{eq:posterior_feature}
\end{equation}
Here, $\rho$ uses two $3^3$ convolutions with group normalization and GELU, then predicts gate $\mathbf{r}=\sigma(W_r\mathbf{h})$ and class residual $\boldsymbol{\Delta}=W_{\Delta}\mathbf{h}$.
The residual corrects the log posterior before renormalization:
\begin{equation}
\mathbf{p}=
\operatorname{Softmax}
\left(
\log\!\left(\operatorname{clamp}(\mathbf{p}^{\mathrm{base}},\epsilon)\right)
+\mathbf{r}\odot\boldsymbol{\Delta}
\right).
\label{eq:posterior_refinement}
\end{equation}
Zero-initialized $W_{\Delta}$ initially recovers the base prediction; the gate localizes corrections and the residual redistributes class probability.

\subsection{Learning Objective}
\label{subsec:objective}

\paragraph{Multi-level supervision.}
For probability map $\widehat{\mathbf{p}}$ and one-hot target $\mathbf{y}$, let $\mathcal{L}_{\mathrm{seg}}=\mathcal{L}_{\mathrm{WCE}}+\mathcal{L}_{\mathrm{Dice}}$, where the WCE class weight is one minus the current-sample class frequency.
A parameter-shared auxiliary decoder yields four modality-specific predictions $\{\mathbf{p}^{\mathrm{aux}}_m\}_{m=1}^{M}$, and four decoder heads yield deep predictions $\{\mathbf{p}^{\mathrm{ds}}_j\}_{j=1}^{4}$:
\begin{equation}
\begin{aligned}
\mathcal{L}_{\mathrm{multi}}
={}&\mathcal{L}_{\mathrm{seg}}(\mathbf{p},\mathbf{y})\\
&+\sum_{m=1}^{M}
\mathcal{L}_{\mathrm{seg}}(\mathbf{p}^{\mathrm{aux}}_m,\mathbf{y})\\
&+\sum_{j=1}^{4}
\mathcal{L}_{\mathrm{seg}}(\mathbf{p}^{\mathrm{ds}}_j,\mathbf{y}).
\end{aligned}
\label{eq:multi_loss}
\end{equation}
Because training scans contain all four sequences, every modality-specific auxiliary branch receives supervision; the sampled availability mask conditions only the fused path.

\paragraph{ET-aware refinement.}
ET is the smallest region and is especially prone to false negatives without T1ce, so the fused output receives a Tversky loss:
\begin{equation}
\mathcal{L}_{\mathrm{ET}}
=
1-
\frac{\mathrm{TP}_{\mathrm{ET}}+\epsilon}
{\mathrm{TP}_{\mathrm{ET}}
+\alpha\mathrm{FP}_{\mathrm{ET}}
+\beta\mathrm{FN}_{\mathrm{ET}}
+\epsilon}.
\label{eq:et_tversky}
\end{equation}
We use $\alpha=0.3$ and $\beta=0.7$.
The posterior gate receives a soft residual-error target and binary cross-entropy supervision:
\begin{equation}
\begin{aligned}
\mathbf{e}
&=\mathbf{y}_{\mathrm{ET}}\odot
\left[1-\operatorname{sg}(\mathbf{p}^{\mathrm{base}}_{\mathrm{ET}})\right],\\
\mathcal{L}_{\mathrm{gate}}
&=\operatorname{BCE}(\mathbf{r},\mathbf{e}),
\end{aligned}
\label{eq:gate_loss}
\end{equation}
where $\operatorname{sg}(\cdot)$ stops gradients through the base prediction.

\paragraph{Consistency and total objective.}
Every second iteration, the stop-gradient complete-mask prediction $\mathbf{p}^{\mathrm{full}}=f_{\theta}(\mathbf{X},\mathbf{1})$ supervises the incomplete prediction:
\begin{equation}
\mathcal{L}_{\mathrm{con}}
=
D_{\mathrm{KL}}
\left(
\operatorname{sg}(\mathbf{p}^{\mathrm{full}})
\;\middle\|\;
\mathbf{p}
\right).
\label{eq:consistency_loss}
\end{equation}
The final objective is
\begin{equation}
\mathcal{L}_{\mathrm{total}}
=
\mathcal{L}_{\mathrm{multi}}
+\lambda_{\mathrm{ET}}\mathcal{L}_{\mathrm{ET}}
+\lambda_{\mathrm{gate}}\mathcal{L}_{\mathrm{gate}}
+\lambda_{\mathrm{con}}\mathcal{L}_{\mathrm{con}},
\label{eq:objective}
\end{equation}
where $\lambda_{\mathrm{ET}}=\lambda_{\mathrm{gate}}=\lambda_{\mathrm{con}}=0.2$.
Only $\mathbf{p}$ is retained at inference; auxiliary and deep predictions are training-only.

%% file: tables/brats2019_per_config_table.tex
\begin{table*}[!t]
\captionsetup{skip=2pt}
\centering
\begingroup
\small
\setlength{\tabcolsep}{0.75pt}
\renewcommand{\arraystretch}{0.88}
\begin{tabular*}{\textwidth}{@{\extracolsep{\fill}}c l
>{\columncolor{configband}}c c
>{\columncolor{configband}}c c
>{\columncolor{configband}}c c
>{\columncolor{configband}}c c
>{\columncolor{configband}}c c
>{\columncolor{configband}}c c
>{\columncolor{configband}}c c
>{\columncolor{configband}}c c@{}}
\toprule
& \textbf{Model}
& \AvailHeader{1}{0}{0}{0}
& \AvailHeader{0}{1}{0}{0}
& \AvailHeader{0}{0}{1}{0}
& \AvailHeader{0}{0}{0}{1}
& \AvailHeader{1}{1}{0}{0}
& \AvailHeader{1}{0}{1}{0}
& \AvailHeader{1}{0}{0}{1}
& \AvailHeader{0}{1}{1}{0}
& \AvailHeader{0}{1}{0}{1}
& \AvailHeader{0}{0}{1}{1}
& \AvailHeader{1}{1}{1}{0}
& \AvailHeader{1}{1}{0}{1}
& \AvailHeader{1}{0}{1}{1}
& \AvailHeader{0}{1}{1}{1}
& \AvailHeader{1}{1}{1}{1}
& \cellcolor{white}\textbf{Avg.} \\
\midrule
\input{tables/brats2019_per_config_rows.tex}\end{tabular*}
\endgroup
\caption{DSC\% ($\uparrow$) comparison across different missing modalities on BraTS2019. Filled and dashed circles denote present and missing modalities, respectively. Our proposal is identified with $\dagger$. Best and second best results in \textbf{bold} and \underline{underline}, respectively. $^{*}$ denotes a statistically significant improvement over the corresponding baseline across the 15 missing-modality configurations using a paired Wilcoxon signed-rank test with Holm correction ($p<0.05$).}
\label{tab:brats2019}

\end{table*}

%% file: tables/brats2019_per_config_rows.tex
\multirow{7}{*}{\textbf{ET}}
& U-HVED
& 19.7 & 16.7 & 49.1 & 4.3 & 31.2 & 61.9 & 12.6 & 59.6
& 17.3 & 52.5 & 64.0 & 19.4 & 62.5 & 57.7 & 62.8 & 39.4 \\
& RobustSeg
& 31.7 & 23.3 & 65.2 & 34.2 & 37.8 & 69.9 & 41.2 & 69.1
& 38.5 & 69.7 & 71.6 & 41.9 & 70.0 & 71.4 & 71.2 & 53.8 \\
& SFusion
& \textbf{68.5} & 39.0 & 29.6 & 34.6 & \textbf{71.0} & 70.6 & \textbf{70.5} & 39.5
& 42.7 & 40.1 & 71.0 & \textbf{70.6} & 71.0 & 44.2 & 70.2 & 55.5 \\
& mmFormer
& 40.7 & \underline{43.4} & \underline{72.6} & \underline{42.4} & 47.0 & 74.1 & 46.1 & \underline{73.8}
& 45.9 & \underline{74.9} & \underline{74.6} & 48.2 & 74.6 & 73.3 & \textbf{76.3} & \underline{60.5} \\
& M$^3$AE
& 38.2 & 29.3 & 71.2 & 35.4 & 38.5 & \underline{75.3} & 40.6 & 72.3
& 36.8 & 74.6 & \textbf{75.8} & 42.5 & \textbf{76.6} & \underline{75.0} & 75.0 & 57.1 \\
& IM-Fuse
& 42.4 & 40.3 & 71.5 & 42.3 & \underline{48.1} & 72.1 & 48.5 & 72.1
& \underline{46.7} & 72.7 & 73.6 & \underline{50.5} & 72.2 & 72.8 & 72.2 & 59.9 \\
& \textbf{Ours}$^{\dagger}$
& \underline{43.4} & \textbf{45.4} & \textbf{75.6} & \textbf{44.9} & 47.6 & \textbf{75.5} & \underline{49.2} & \textbf{75.2}
& \textbf{46.8} & \textbf{75.4} & 73.9 & 49.3 & \underline{74.9} & \textbf{76.0} & \underline{75.2} & \textbf{61.9}$^{*}$ \\
\midrule
\multirow{7}{*}{\textbf{TC}}
& U-HVED
& 40.2 & 45.5 & 50.6 & 47.7 & 58.0 & 63.2 & 47.3 & 65.4
& 51.8 & 57.8 & 70.8 & 53.7 & 65.3 & 68.7 & 72.5 & 57.2 \\
& RobustSeg
& 62.1 & 51.8 & 74.0 & 64.2 & 68.8 & 81.2 & 70.7 & 79.0
& 68.9 & 81.5 & 82.1 & 72.0 & 82.4 & 82.6 & 83.2 & 73.6 \\
& SFusion
& \textbf{80.2} & 67.7 & 60.8 & 63.6 & \textbf{83.7} & 82.1 & \textbf{83.8} & 70.8
& 72.2 & 70.4 & 84.5 & \textbf{84.4} & 84.9 & 74.2 & 84.7 & 76.5 \\
& mmFormer
& 70.8 & \underline{70.9} & \underline{83.1} & \textbf{72.4} & 76.1 & \underline{84.7} & 76.5 & \underline{84.4}
& \underline{74.7} & \underline{84.6} & \underline{85.0} & 77.9 & \textbf{85.5} & 84.9 & \underline{85.3} & \underline{79.8} \\
& M$^3$AE
& 62.6 & 55.8 & 77.0 & 62.3 & 67.0 & 81.6 & 66.8 & 80.8
& 65.1 & 79.0 & 83.1 & 68.2 & 81.9 & 80.9 & 82.3 & 73.0 \\
& IM-Fuse
& 73.2 & 70.6 & 82.7 & 71.6 & 77.4 & 84.4 & 76.7 & 83.8
& 74.1 & 83.6 & 84.8 & 77.5 & 84.2 & \underline{84.2} & 84.4 & 79.5 \\
& \textbf{Ours}$^{\dagger}$
& \underline{74.3} & \textbf{71.4} & \textbf{83.5} & \underline{71.8} & \underline{78.1} & \textbf{85.1} & \underline{77.5} & \textbf{84.7}
& \textbf{75.5} & \textbf{84.7} & \textbf{85.6} & \underline{78.2} & \underline{85.1} & \textbf{84.9} & \textbf{85.4} & \textbf{80.4}$^{*}$ \\
\midrule
\multirow{7}{*}{\textbf{WT}}
& U-HVED
& 44.3 & \underline{79.6} & 45.4 & 76.1 & 79.8 & 58.4 & 77.9 & 78.0
& 84.7 & 77.5 & 79.4 & 84.1 & 79.7 & 83.5 & 83.8 & 74.1 \\
& RobustSeg
& 83.7 & 65.3 & 68.9 & 83.7 & 87.4 & 86.2 & 87.8 & 74.2
& 85.1 & 85.3 & 87.4 & 88.5 & 88.3 & 85.8 & 88.4 & 83.1 \\
& SFusion
& 71.4 & \textbf{82.0} & 68.5 & 84.9 & 85.1 & 75.5 & 88.7 & \textbf{83.6}
& 88.6 & \textbf{88.5} & 85.6 & 89.7 & 89.8 & \textbf{89.5} & 90.3 & 84.1 \\
& mmFormer
& \underline{88.5} & 77.3 & \underline{78.2} & 86.4 & \underline{90.3} & \textbf{90.5} & \underline{90.3} & 80.8
& 88.1 & 88.4 & \textbf{90.9} & \underline{90.9} & \underline{91.1} & 88.6 & \underline{91.4} & \underline{87.4} \\
& M$^3$AE
& 85.0 & 69.7 & 70.7 & 82.0 & 88.5 & 87.2 & 88.7 & 73.4
& 85.3 & 83.3 & 88.5 & 89.5 & 88.2 & 85.0 & 89.6 & 83.6 \\
& IM-Fuse
& 87.8 & 77.2 & 78.1 & \underline{86.8} & 89.5 & 89.3 & 90.2 & 80.8
& \underline{88.6} & 88.4 & 90.3 & 90.8 & 90.6 & 88.3 & 91.2 & 87.2 \\
& \textbf{Ours}$^{\dagger}$
& \textbf{88.9} & 77.9 & \textbf{79.7} & \textbf{87.8} & \textbf{90.7} & \underline{90.2} & \textbf{90.8} & \underline{81.3}
& \textbf{89.2} & \underline{88.4} & \underline{90.6} & \textbf{91.4} & \textbf{91.3} & \underline{88.6} & \textbf{91.5} & \textbf{87.9}$^{*}$ \\
\bottomrule

%% file: tables/brats2023_per_config_table.tex
\begin{table*}[!t]
\captionsetup{skip=2pt}
\centering
\begingroup
\small
\setlength{\tabcolsep}{0.75pt}
\renewcommand{\arraystretch}{0.88}
\begin{tabular*}{\textwidth}{@{\extracolsep{\fill}}c l
>{\columncolor{configband}}c c
>{\columncolor{configband}}c c
>{\columncolor{configband}}c c
>{\columncolor{configband}}c c
>{\columncolor{configband}}c c
>{\columncolor{configband}}c c
>{\columncolor{configband}}c c
>{\columncolor{configband}}c c@{}}
\toprule
& \textbf{Model}
& \AvailHeader{1}{0}{0}{0}
& \AvailHeader{0}{1}{0}{0}
& \AvailHeader{0}{0}{1}{0}
& \AvailHeader{0}{0}{0}{1}
& \AvailHeader{1}{1}{0}{0}
& \AvailHeader{1}{0}{1}{0}
& \AvailHeader{1}{0}{0}{1}
& \AvailHeader{0}{1}{1}{0}
& \AvailHeader{0}{1}{0}{1}
& \AvailHeader{0}{0}{1}{1}
& \AvailHeader{1}{1}{1}{0}
& \AvailHeader{1}{1}{0}{1}
& \AvailHeader{1}{0}{1}{1}
& \AvailHeader{0}{1}{1}{1}
& \AvailHeader{1}{1}{1}{1}
& \cellcolor{white}\textbf{Avg.} \\
\midrule
\input{tables/brats2023_per_config_rows.tex}\end{tabular*}
\endgroup
\caption{DSC\% ($\uparrow$) comparison across different missing modalities on BraTS2023. Filled and dashed circles denote present and missing modalities, respectively. Our proposal is identified with $\dagger$. Best and second best results in \textbf{bold} and \underline{underline}, respectively. $^{*}$ denotes a statistically significant improvement over the corresponding baseline across the 15 missing-modality configurations using a paired Wilcoxon signed-rank test with Holm correction ($p<0.05$).}
\label{tab:brats2023}

\end{table*}

%% file: tables/brats2023_per_config_rows.tex
\multirow{9}{*}{\textbf{ET}}
& U-HVED
& 39.9 & 21.5 & 76.6 & 40.1 & 43.9 & 76.5 & 47.5 & 73.3
& 36.1 & 78.2 & 78.5 & 47.6 & 78.2 & 79.2 & 79.3 & 59.8 \\
& RobustSeg
& 50.3 & 44.8 & 80.2 & 50.5 & 55.4 & 82.7 & 57.4 & 81.7
& 53.9 & 82.4 & 83.7 & 59.0 & 82.8 & 82.9 & 83.6 & 68.8 \\
& SFusion
& 52.2 & 48.9 & 82.2 & 54.3 & 57.4 & 83.9 & 59.0 & 83.6
& 57.1 & 83.8 & 83.8 & 60.1 & 83.9 & 84.0 & 84.0 & 70.6 \\
& mmFormer
& 58.8 & 54.7 & \underline{84.1} & 58.3 & 62.5 & 84.9 & 63.9 & \underline{84.7}
& \underline{62.5} & 84.7 & 84.8 & \underline{66.0} & 84.2 & \textbf{85.9} & 84.7 & 73.6 \\
& ShaSpec
& 53.3 & 49.1 & 80.5 & 52.4 & 57.4 & 81.9 & 58.0 & 81.7
& 56.1 & 81.9 & 82.4 & 59.6 & 82.1 & 82.4 & 82.4 & 69.4 \\
& M$^3$AE
& 56.7 & 56.0 & 82.5 & 58.8 & 60.7 & \textbf{85.6} & 61.2 & \textbf{84.9}
& 60.6 & \textbf{85.8} & \textbf{85.9} & 62.3 & \textbf{85.7} & 85.1 & \underline{85.6} & 73.2 \\
& M$^3$FeCon
& 53.2 & 53.8 & 82.8 & 56.6 & 58.0 & 83.5 & 60.4 & 84.3
& 61.2 & 83.9 & 84.0 & 62.4 & 84.0 & 84.4 & 84.2 & 71.8 \\
& IM-Fuse
& \underline{59.5} & \underline{56.3} & 83.5 & \textbf{59.6} & \underline{63.9} & \underline{85.0} & \textbf{64.3} & 84.5
& \textbf{63.6} & 84.9 & \underline{85.1} & \textbf{67.0} & \underline{85.2} & \underline{85.6} & \textbf{85.8} & \textbf{74.3} \\
& \textbf{Ours}$^{\dagger}$
& \textbf{59.7} & \textbf{56.8} & \textbf{84.5} & \underline{58.9} & \textbf{64.0} & 84.0 & \underline{64.0} & 84.2
& 62.0 & \underline{85.1} & 83.8 & 65.1 & 84.0 & 84.0 & 83.6 & \underline{73.6}$^{*}$ \\
\midrule
\multirow{9}{*}{\textbf{TC}}
& U-HVED
& 58.9 & 47.1 & 81.9 & 61.7 & 68.0 & 81.9 & 66.5 & 84.9
& 60.2 & 84.8 & 84.4 & 68.9 & 84.4 & 85.3 & 86.0 & 73.7 \\
& RobustSeg
& 70.2 & 66.9 & 85.9 & 69.8 & 76.3 & 89.6 & 76.0 & 87.7
& 73.9 & 88.4 & 90.3 & 78.1 & 90.0 & 89.6 & 90.5 & 81.5 \\
& SFusion
& 74.0 & 70.4 & 86.7 & 74.3 & 77.4 & 88.6 & 77.8 & 88.4
& 76.5 & 89.4 & 89.1 & 78.5 & 89.3 & 89.5 & 89.5 & 82.6 \\
& mmFormer
& 78.3 & 73.6 & 89.2 & 74.5 & 80.6 & 90.7 & \underline{80.0} & 90.1
& 77.5 & 90.6 & 90.9 & 80.8 & 91.0 & 90.8 & 91.0 & 84.7 \\
& ShaSpec
& 74.3 & 71.5 & 87.8 & 72.6 & 78.1 & 89.7 & 77.3 & 89.3
& 76.2 & 89.6 & 90.3 & 79.1 & 90.4 & 90.0 & 90.7 & 83.1 \\
& M$^3$AE
& 76.8 & \underline{75.9} & 89.9 & \textbf{77.9} & 79.9 & \underline{90.9} & 79.2 & 90.5
& 78.9 & \underline{90.8} & \underline{91.2} & 79.9 & \textbf{91.5} & 91.0 & 91.5 & \underline{85.1} \\
& M$^3$FeCon
& 72.3 & 74.1 & \underline{90.1} & 75.9 & 77.1 & 90.9 & 79.1 & \underline{90.6}
& \underline{79.5} & 90.7 & 91.2 & 80.9 & 91.2 & \underline{91.4} & \underline{91.1} & 84.4 \\
& IM-Fuse
& \underline{78.8} & 75.4 & \textbf{90.5} & 76.5 & \underline{80.9} & \textbf{91.4} & 79.9 & \textbf{91.2}
& 79.1 & \textbf{91.2} & \textbf{91.6} & \underline{81.4} & \underline{91.3} & \textbf{91.5} & \textbf{91.5} & 85.5 \\
& \textbf{Ours}$^{\dagger}$
& \textbf{79.9} & \textbf{76.7} & 89.5 & \underline{77.7} & \textbf{81.8} & 90.5 & \textbf{81.5} & 90.2
& \textbf{79.8} & 90.2 & 90.6 & \textbf{82.1} & 90.5 & 90.4 & 90.6 & \textbf{85.5}$^{*}$ \\
\midrule
\multirow{9}{*}{\textbf{WT}}
& U-HVED
& 82.8 & 71.2 & 63.5 & 80.1 & 86.8 & 86.8 & 87.9 & 79.0
& 83.9 & 85.1 & 88.4 & 89.3 & 90.0 & 86.8 & 90.8 & 83.5 \\
& RobustSeg
& 88.7 & 74.7 & 76.1 & 85.3 & 90.6 & 91.2 & 90.8 & 79.7
& 87.3 & 89.3 & 91.6 & 91.4 & 92.0 & 88.7 & 92.2 & 87.3 \\
& SFusion
& 89.1 & 78.5 & 77.6 & 87.0 & 90.8 & 91.2 & 91.3 & 81.3
& 88.2 & 88.7 & 91.7 & 91.6 & 92.1 & 88.8 & 92.2 & 88.0 \\
& mmFormer
& 91.4 & 82.8 & 83.7 & 88.5 & 92.2 & 92.7 & 91.3 & 85.5
& \underline{89.8} & 90.1 & 92.8 & 92.5 & 93.0 & 90.5 & 93.0 & 90.0 \\
& ShaSpec
& 91.0 & 79.9 & 79.5 & 86.9 & 91.9 & 92.3 & 92.2 & 82.7
& 88.3 & 88.7 & 92.6 & 92.5 & 92.9 & 89.2 & 93.0 & 88.8 \\
& M$^3$AE
& 91.5 & 81.7 & 82.5 & 88.5 & 91.9 & 92.5 & 92.2 & 83.6
& 89.2 & 89.7 & 92.6 & 92.1 & 93.0 & 90.0 & 92.9 & 89.6 \\
& M$^3$FeCon
& 87.7 & 81.2 & 81.2 & 88.5 & 89.4 & 90.0 & 92.1 & 83.9
& 89.8 & 89.5 & 90.5 & \underline{92.7} & 92.6 & 90.1 & \underline{93.1} & 88.8 \\
& IM-Fuse
& \underline{91.8} & \underline{83.0} & \underline{83.7} & \underline{88.7} & \underline{92.4} & \underline{92.8} & \underline{92.6} & \underline{85.5}
& 90.1 & \underline{90.2} & \underline{92.8} & 93.0 & \underline{93.1} & \underline{90.5} & 93.3 & \underline{90.2} \\
& \textbf{Ours}$^{\dagger}$
& \textbf{91.9} & \textbf{83.8} & \textbf{83.8} & \textbf{89.0} & \textbf{92.6} & \textbf{93.0} & \textbf{92.8} & \textbf{85.7}
& \textbf{90.1} & \textbf{90.5} & \textbf{93.1} & \textbf{93.0} & \textbf{93.3} & \textbf{90.8} & \textbf{93.3} & \textbf{90.5}$^{*}$ \\
\bottomrule

%% file: tables/fusion_ablation_table.tex
\begin{table}[!b]
\centering
\begingroup
\small
\setlength{\tabcolsep}{1.4pt}
\renewcommand{\arraystretch}{1.05}
\begin{tabular}{@{}
>{\raggedright\arraybackslash}p{0.13\columnwidth}
>{\raggedright\arraybackslash}p{0.16\columnwidth}
>{\raggedright\arraybackslash}p{0.18\columnwidth}
rrrr@{}}
\toprule
Variant & \shortstack[l]{Lateral\\fusion} & \shortstack[l]{Bottleneck\\fusion} &
WT $\uparrow$ & TC $\uparrow$ & ET $\uparrow$ & \shortstack{Mean\\DSC $\uparrow$} \\
\midrule
Base & Masked mean & Masked mean & 86.21 & 78.29 & 57.01 & 73.84 \\
Uniform ACF & Uniform ACF & Uniform ACF & 85.02 & 77.54 & 56.63 & 73.06 \\
w/o HACF & Masked mean & TriRSF & 85.89 & 78.12 & 57.27 & 73.76 \\
w/o TriRSF & HACF & Masked mean & 85.77 & 77.82 & 56.70 & 73.43 \\
Full & HACF & TriRSF & \textbf{87.89} & \textbf{80.38} & \textbf{61.88} & \textbf{76.72} \\
\bottomrule
\end{tabular}
\endgroup
\caption{Ablation study of the proposed hierarchical fusion framework under all missing-modality scenarios on BraTS2019. Different variants evaluate the effectiveness of HACF and TriRSF at lateral and bottleneck feature levels. Results are reported using the DSC (\%) metric for WT, TC, ET, and Mean DSC.}
\label{tab:ablation}
\end{table}